\documentclass[letterpaper,10pt,conference]{ieeeconf}
\IEEEoverridecommandlockouts
\usepackage{makecell}
\usepackage[T1]{fontenc}
\usepackage{cite}
\usepackage{amsmath,amssymb,amsfonts}
\usepackage{algorithmic}
\usepackage{algorithm}
\usepackage{graphicx}
\usepackage{booktabs}
\usepackage{multirow}
\usepackage{tabularx}
\usepackage{array}
\usepackage{makecell}
\usepackage{textcomp}
\usepackage{xcolor}
\usepackage{hyperref}
\usepackage{microtype}
\begin{document}

\title{Closed-Loop Control with Rule-Aligned Small Language Models and Multi-Agent Self-Correction\\
}

\author{
\begin{tabular}{cc}
\begin{tabular}[t]{@{}c@{}}
\textbf{Yuchen Wang}\\
\textit{School of Computer Science}\\
\textit{University of Sheffield}\\
Sheffield, UK\\
ywang1016@sheffield.ac.uk
\end{tabular}
&
\begin{tabular}[t]{@{}c@{}}
\textbf{Javal Vyas}\\
\textit{Department of Chemical Engineering}\\
\textit{Imperial College London}\\
London, UK\\
j.vyas24@imperial.ac.uk
\end{tabular}
\\[1.2em]
\begin{tabular}[t]{@{}c@{}}
\textbf{Tong Liu}\\
\textit{School of Computer Science}\\
\textit{University of Sheffield}\\
Sheffield, UK\\
t.liu@sheffield.ac.uk
\end{tabular}
&
\begin{tabular}[t]{@{}c@{}}
\textbf{Mehmet Mercang{\"o}z}\\
\textit{Department of Chemical Engineering}\\
\textit{Imperial College London}\\
London, UK\\
m.mercangoz@imperial.ac.uk
\end{tabular}
\end{tabular}
}

\maketitle

\begin{abstract}

A key step toward autonomous industrial operation is the ability to create and reconfigure control policies from natural-language requirement specifications, with minimal or no manual redesign. In this setting, policy generation by AI agents can be a credible path when paired with a plant-aware validator (e.g., a digital twin) that can check generated candidate actions before execution. However, practical deployment is constrained by inference latency and compute footprint: large cloud-based models are often too slow, opaque, or data-sensitive for edge closed-loop use. This work investigates whether a compact Small Language Model (SLM) can be retrained for control reasoning and embedded in a validator-guided correction loop. We use a Qwen2.5-1.5B model aligned via Group Relative Policy Optimization (GRPO), combined with (i) an action agent, (ii) a symbolic/digital-twin-style validation layer, and (iii) a reprompting agent that iteratively steers outputs toward valid actions. In randomized thermal-control simulations (30 experiments with 500 steps each), the framework achieves 91.5\% average action-alignment accuracy (86.3\%--100\% across cases) at 3.84\,s mean inference latency. Under symbolic re-mapping, it maintains a 95\% in-range rate, indicating robust physical regulation despite reduced token-level agreement. These results support SLM+validator architectures as a practical path toward reconfigurable autonomous control at the edge.
\end{abstract}

\textbf{Keywords:}
Autonomous systems, Small language models, Edge AI, Generative AI

\section{Introduction}

A central requirement for autonomous industrial operation is the ability to \emph{create and reconfigure control policies} from high-level, natural-language requirement specifications with minimal manual redesign. This is particularly important in environments where objectives and constraints evolve over time (e.g., energy priorities, safety envelopes, and operating policies).  In this setting, policy generation by AI agents can be a credible path when paired with a plant-aware validator (e.g., a digital twin) that can check generated candidate actions before execution \cite{11205597}.

Conventional industrial control architectures are highly effective for deterministic numerical regulation, but they are less flexible when symbolic operating intent must be incorporated quickly. Translating new expert directives into deployed logic typically requires manual re-engineering of rules, supervisory layers, or interlocks, which slows reconfiguration and increases engineering overhead. This creates a persistent gap between high-level operational intent and low-level control execution.

Language models provide a promising interface for closing this gap because they can parse and operationalize natural-language instructions and support iterative self-correction during inference \cite{wang2023voyager, wei2022chain}. However, deploying cloud-hosted Large Language Models (LLMs) in closed-loop industrial workflows remains challenging: API dependence introduces variable end-to-end latency, while external data transfer raises governance and sovereignty concerns for sensitive operational telemetry \cite{shi2016edge, pedreira2021review}. These issues are especially limiting for edge and offline autonomy.

This work investigates whether a compact Small Language Model (SLM) can be trained to perform control-oriented reasoning and embedded into a validator-guided correction loop. We develop a local architecture built around Qwen2.5-1.5B \cite{hui2024qwen2}, where policy behavior is aligned using Group Relative Policy Optimization (GRPO) \cite{guo2025deepseek}. A key advantage of this RL-based alignment is that the model autonomously generates its own chains-of-thought (CoT). While a brief warm-up distillation is employed, it serves strictly for format alignment rather than task-specific memorization. Furthermore, the framework operates in a zero-shot capacity regarding corrective interactions; although GRPO is utilized solely to enhance the agent’s numerical and logical reasoning, the Action Agent is never exposed to the diverse reprompting hints during training. Nevertheless, it demonstrates an emergent ability to dynamically update its decisions based on such corrective feedback. The controller is implemented as three cooperating components: (i) an Action Agent that proposes control moves, (ii) a symbolic/digital-twin-style validator that checks action validity, and (iii) a Reprompt Agent that provides corrective guidance when validation fails. This design targets a practical balance between reasoning capability, latency, and edge deployability.

\noindent\textbf{Contributions:} The main contributions are threefold: (1) \textit{Rule-aligned SLM policy learning}, providing a pipeline that aligns a 1.5B SLM with thermal control logic using GRPO; (2) a \textit{zero-shot validator-guided inference loop} that enables compact SLMs to exhibit emergent correction capabilities without explicit corrective training; and (3) an \textit{empirical evaluation} demonstrating strong physical regulation (95\% IRR) under interface perturbations in randomized simulations.

\section{Related Work}

\subsection{LLM-driven Autonomous Control and its Advantages}
Traditional automation systems primarily rely on rigid, hard-coded logic to manage industrial processes. While these systems are highly reliable for numerical regulation, they struggle to integrate complex, symbolic expert intents.
Recent research has shifted toward LLM-based Autonomous Agents, where models act as "reasoning engines" to bridge this gap. This paradigm shift has demonstrated significant potential in laboratory discovery \cite{baldea2025automated}, digital workflows \cite{pr13092680}, and robotic reward engineering \cite{ma2023eureka}. 

\subsection{Enhancing Numerical Reasoning in Language Models}
Thermal management requires precise numerical comparison and threshold-based logic, areas where standard LLMs often struggle with "stochastic drift" or calculation errors \cite{shao2024deepseekmath}. Current strategies to mitigate this include:

\textbf{Reinforcement Learning (RL):} Techniques like PPO and particularly Group Relative Policy Optimization (GRPO) \cite{guo2025deepseek, guo2024controlagent} have emerged to align model outputs with verifiable rewards, reinforcing the logical consistency behind numerical decisions.

\textbf{Self-Correction and Recursive Reasoning:} To stabilize control outputs, recent research employs external validation loops and reprompting agents \cite{VYAS2025349, vyas2025autonomous}. However, purely external designs often lack the intrinsic logical depth required for real-time stability. Frameworks like the \textbf{Self-Taught Reasoner (STaR)} \cite{zelikman2022star} address this by enabling models to iteratively refine their own reasoning chains (\textbf{Chain-of-Thought}) based on successful outcomes. This allows the model to internalize complex logic during training—an evolution we extend by using GRPO to foster the "Aha-moment" \cite{guo2025deepseek} within small-scale architectures, ensuring they reason correctly without the overhead of heavy external supervision.

\textbf{Tool Utilization:} Although agents can call external calculators \cite{romero2025agentic,hu2025autocontrol}, the resulting communication latency and instability make this approach not suitable for all problems.

\subsection{Small Language Models (SLMs) for Edge Deployment}
While Large Language Models (LLMs) possess superior reasoning, their prohibitive inference latency and deployment costs hinder real-time industrial integration. Small Language Models (SLMs), when properly optimized, are being explored to achieve functional performance in specialized tasks while offering the potential for local execution in thermal control \cite{belcak2025small}. Recent studies emphasize using SLMs as "control-specific" agents that internalize expert experience to replace massive general-purpose architectures \cite{guo2024controlagent}. This paper extends this direction by investigating the feasibility of addressing the reasoning gap in 1.5B-parameter SLMs through GRPO-based heuristic internalization.

\section{Methodology}

Our architecture employs a multi-agent collaboration to ensure system's self-correction.

\subsection{Collaborative Multi-Agent Architecture}
The core of the closed-loop system consists of three specialized agents that partition the tasks of decision-making, verification, and logical correction.

    
    

 \textbf{Action Agent (Reasoning Core):} Powered by Qwen2.5-1.5B and aligned via GRPO, this agent processes the state $s_t$ and context to generate a control action $u_t$ alongside a Chain-of-Thought (CoT) reasoning path.

\textbf{Validation Agent (Symbolic Guardrail):} Enforcing a strict logical filter, this component flags the output $u_t$ as \textit{UNSAFE} if it diverges from the ground-truth expert rule $u_{gt}$.

\textbf{Reprompt Agent (Logical Interpreter):} Also utilizing Qwen2.5-1.5B, this agent diagnoses discrepancies between $u_t$ and $u_{gt}$. Fine-tuned on the Action Agent's error distributions, it constructs a semantic \textit{Corrective Hint} $h_t$ regarding boundary or rule oversights to guide zero-shot self-correction. \looseness=-1

\subsection{The Control Algorithm}
\begin{figure}
    \centering
    \includegraphics[width=0.85\linewidth]{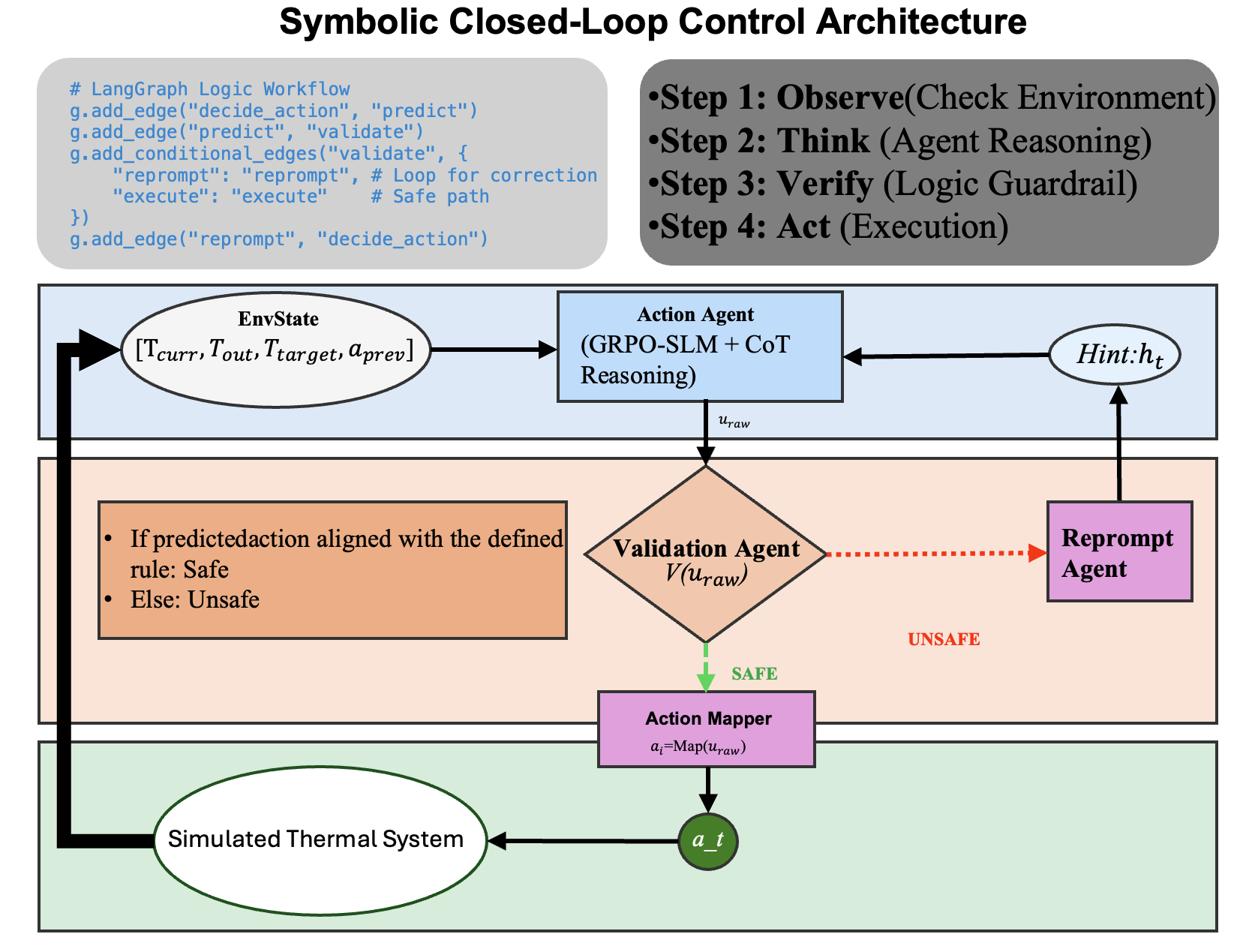} 
    \caption{Schematic of the Agentic Closed-Loop Control Architecture. The framework incorporates a GRPO-aligned Small Language Model (SLM) for decision reasoning (blue) and a Symbolic Validation (orange) for rule checking. The Execution layer(green) is to update the system continuously after receiving one valid action.}
    \label{fig:framework}
\end{figure}

The interaction between these agents is formalized in Algorithm~\ref{alg:agent_loop} and visualized in the system architecture shown in Fig.~\ref{fig:framework}. As illustrated, the framework is structured into three functionally distinct layers: the \textit{Reasoning Layer} (blue), the \textit{Symbolic Validation Layer} (orange), and the \textit{Execution Layer} (green).

\begin{algorithm}[t]
\caption{Multi-Agent Collaborative Control Loop}
\label{alg:agent_loop}
\begin{algorithmic}[1]
\STATE \textbf{Input:} Current system state $s_t$, Maximum retry limit $N_{max}$
\STATE \textbf{Output:} Validated control action $u_t$
\STATE $attempt \leftarrow 0$
\STATE $hint \leftarrow \emptyset$
\WHILE{$attempt < N_{max}$}
    \STATE $u_t, \text{CoT}_t \leftarrow \text{ActionAgent}(s_t, hint)$
    \STATE $u_{gt} \leftarrow \text{ExpertRules}(s_t)$
    \IF{$u_t = u_{gt}$}
        \STATE \RETURN $u_t$
    \ELSE
        \STATE $hint \leftarrow \text{RepromptAgent}(u_t, u_{gt})$
        \STATE $attempt \leftarrow attempt + 1$
    \ENDIF
\ENDWHILE
\STATE \RETURN $u_{gt}$
\end{algorithmic}
\vspace{-0.4em}
\end{algorithm}

\subsection{Control Logic and System Design}
The symbolic rules embedded within the framework are engineered to emulate an expert thermal engineer's heuristics, focusing on three operational pillars:

\subsubsection{\textbf{Control Logic Design}}
The symbolic rules in Table~\ref{table:rules} are engineered to emulate the decision-making process of a thermal system. Unlike reactive systems, our framework incorporates \textbf{temporal state persistence}: the agent's current decision is conditioned not only on the thermal state $S$ but also on the \textbf{memory of the previous action} ($A_{prev}$). This "Latching" mechanism (P3) ensures mission continuity and prevents oscillatory switching during thermal transitions, maintaining stability until the setpoint is reached.

\begin{table}[htbp]
\caption{Symbolic Rule Hierarchy for Thermal Control}
\begin{center}
\resizebox{0.95\columnwidth}{!}{
\begin{tabular}{|l|l|c|}
\hline
\textbf{Priority} & \textbf{Condition (Current State $S$)} & \textbf{Action} \\
\hline
\multicolumn{3}{|c|}{\textit{1. Safety Override (Critical, P1)}} \\
\hline
Safety H & $T_{cur} \geq T_{safe,high}$ & $-2Q$ \\
Safety L & $T_{cur} \leq T_{safe,low}$ & $+2Q$ \\
\hline
\multicolumn{3}{|c|}{\textit{2. Boundary Trigger (Tracking, P2)}} \\
\hline
Upper B & $T_{cur} > T_{target,high}$ & $-Q$ \\
Lower B & $T_{cur} < T_{target,low}$ & $+Q$ \\
\hline
\multicolumn{3}{|c|}{\textit{3. Mission Latching (Hysteresis, P3)}} \\
\hline
Maint. H & $T_{cur} \in [T_{low}, T_{high}] \cap A_{prev} > 0$ & $+Q$ \\
Maint. C & $T_{cur} \in [T_{low}, T_{high}] \cap A_{prev} < 0$ & $-Q$ \\
\hline
\textbf{Default} & All other applicable cases & $0$ \\
\hline
\end{tabular}
}
\label{table:rules}
\end{center}
\vspace{-1em} %
\end{table}
\subsubsection{\textbf{Closed-Loop Reasoning and Validation}}
The control loop is organised as a \emph{verify-then-execute} pipeline (Algorithm~\ref{alg:agent_loop}). At each step, a candidate action is first proposed by the reasoning agent and then screened by a validation layer before it is applied to the plant.

\subsubsection{System Dynamics and Environment}
The thermal environment is modeled via a first-order energy balance equation:
\begin{equation}
C \frac{dT_t}{dt} = UA(T_{\text{out}} - T_t) + a_t + \epsilon
\end{equation}
where $C$ is the room thermal capacity, $UA$ is the overall heat-transfer
coefficient, and $T_{\text{out}}$ is the ambient outdoor temperature. The
control action is selected from the discrete set
$a_t \in \{-2Q, -Q, 0, Q, 2Q\}$, where $Q=5.0$ is the base power unit.

\subsection{ALGORITHM REFINEMENT: GRPO AND DISTILLATION}
To optimize the SLM's reasoning capabilities, we employ a strategy of Group Relative Policy Optimization (GRPO) followed by Knowledge Distillation.

\subsubsection{GRPO and Reward Function}
Unlike PPO which requires a value function $V(s)$, we adopt GRPO to stabilize training via group-based relative advantages \cite{guo2025deepseek}. For a given state $s_t$, we sample $G$ reasoning trajectories $\{y_1, y_2, \dots, y_G\}$. The reward function $R$ is formulated to align the model with symbolic logic. For each trajectory $i$, a reward $r_i$ is assigned based on the logic detailed in Table \ref{tab:reward_logic}.

\begin{table}[htbp]
\centering
\caption{Reward logic and penalty mapping for GRPO training}
\label{tab:reward_logic}

\scriptsize
\setlength{\tabcolsep}{2.5pt}
\renewcommand{\arraystretch}{1.08}

\begin{tabularx}{\columnwidth}{@{}
  >{\raggedright\arraybackslash}X
  >{\centering\arraybackslash}p{0.22\columnwidth}
  >{\centering\arraybackslash}p{0.24\columnwidth}
  >{\centering\arraybackslash}p{0.11\columnwidth}
@{}}
\toprule
\textbf{Condition} &
\textbf{Ground truth} &
\textbf{Output} &
\textbf{Score} \\
\midrule

\multicolumn{4}{@{}c@{}}{\textit{Task persistence (latching)}} \\
\midrule

$a_{t-1} \in \{+Q, +2Q\}^{\mathrm{a}}$ &
$+Q$ &
$0$ (early exit) &
$-0.5$ \\

$a_{t-1} \in \{-Q, -2Q\}^{\mathrm{a}}$ &
$-Q$ &
$0$ (early exit) &
$-0.5$ \\

\midrule
\multicolumn{4}{@{}c@{}}{\textit{Critical errors}} \\
\midrule

Polarity inversion &
e.g., $+Q$ &
e.g., $-Q$ &
$-0.75$ \\

Invalid format &
Any &
Unknown &
$-1.0$ \\

\midrule

\textbf{Correct execution} &
$a^*$ &
$\hat{a}=a^*$ &
$1.0$ \\

\bottomrule
\end{tabularx}

\vspace{-0.8em}
\end{table}

The optimization follows the GRPO algorithm \cite{liu2024deepseek}. For each state $s_t$, the SLM generates a group of $G=8$ independent reasoning trajectories. 
\subsubsection{Action Agent Initialization via Distillation}
The overall training and alignment workflow is illustrated in Fig. \ref{fig:distillation}. To prevent divergence during the subsequent reinforcement learning phase, the \textbf{Action Agent} undergoes an initial brief distillation stage using \textit{DeepSeek-R1-671B} as the teacher. Following the \textit{distilling step-by-step} paradigm \cite{hsieh2023distilling}, the agent learns to internalize the structured reasoning paths (CoT) alongside the control labels using a subset of 5,000 trajectories. This stage provides a necessary ``warm-start'' policy, ensuring that the agent maintains a structured format before being optimized for numerical precision via GRPO.

\subsubsection{Iterative Error-Driven Distillation}
\begin{figure}
    \centering
    \includegraphics[width=0.85\linewidth]{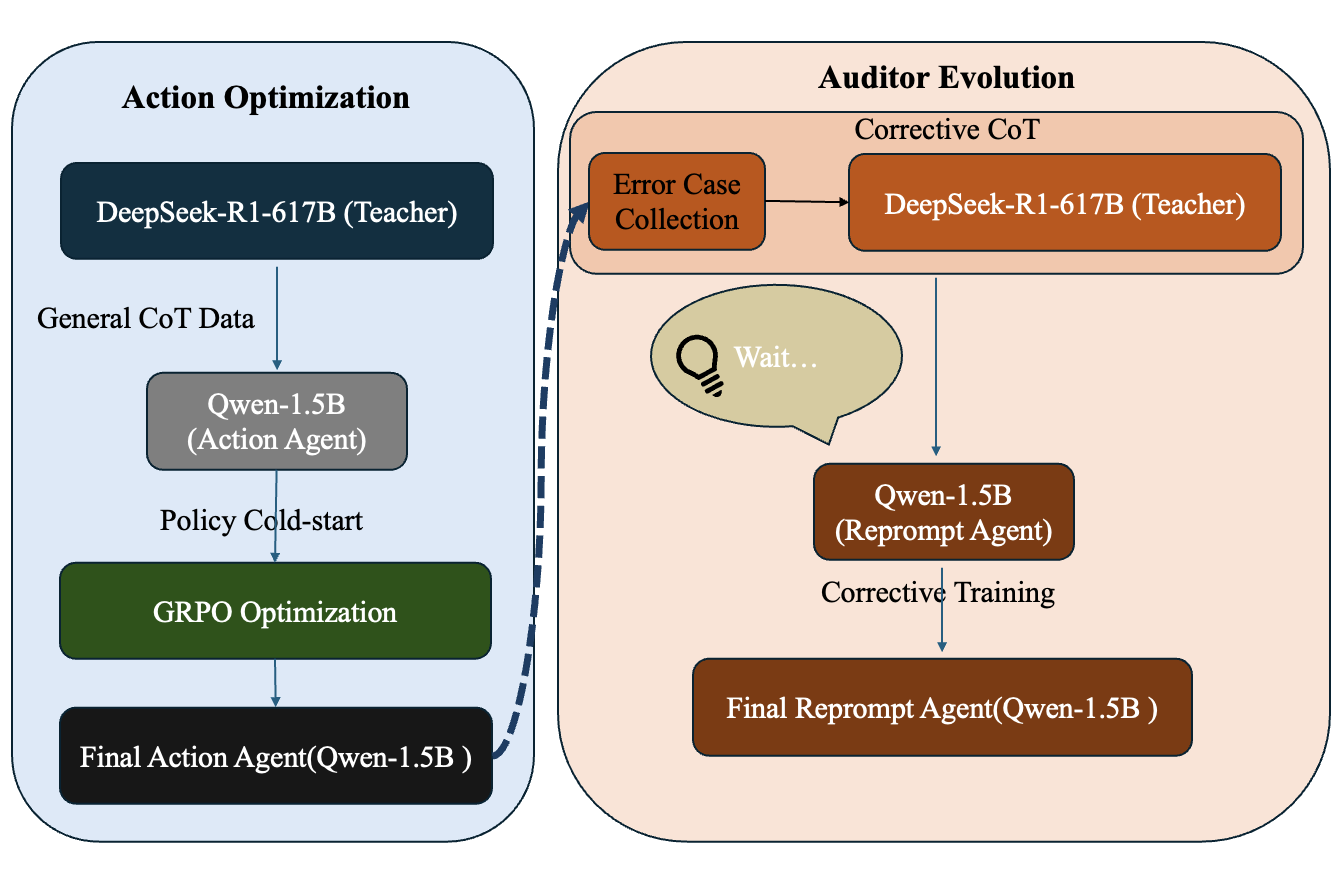}
    \caption{The bifurcated training and self-correction pipeline, featuring a GRPO-based action optimization path (left) and an iterative auditor evolution path (right).}
    \label{fig:distillation}

\end{figure}

Inspired by the \textit{STaR} (Self-Taught Reasoner) framework \cite{zelikman2022star}, we develop an iterative distillation pipeline for the \textbf{Reprompt Agent}, forming the feedback loop shown in the right panel of Fig. \ref{fig:distillation}. The process is executed in three stages:

\noindent\textbf{1) Error Profiling:} After the Action Agent completes its GRPO optimization, we deploy it in challenging thermal environments to collect corner-case failure trajectories. 
\noindent\textbf{2) Teacher Correction:} These error cases are fed into the DeepSeek-R1-671B teacher model to generate 5,000 corrective samples. 
\noindent\textbf{3) Aha-Moment Distillation:} Drawing inspiration from the ``Aha moment,'' the Reprompt Agent is fine-tuned to initiate feedback with a ``Wait...'' prefix, forcing a secondary verification.

\subsection{Prompt Design for Symbolic Reasoning}
\begin{figure}[htbp]
    \centering
    \includegraphics[width=0.85\linewidth]{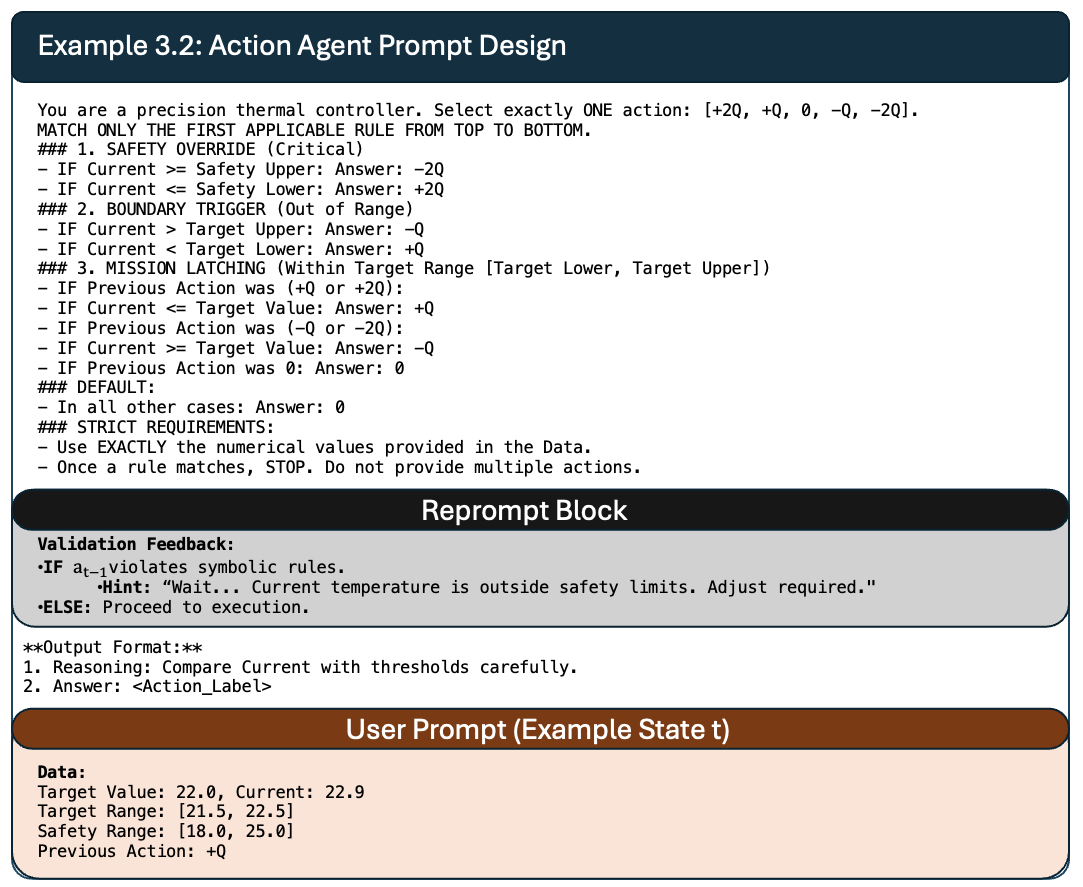}
    \caption{Hierarchical Prompt Design with Symbolic Latching and Reprompt Logic. The structured instruction set ensures deterministic decision-making through prioritized rule matching.}
    \label{fig:prompt_design}

\end{figure}

\begin{figure}[htbp]
    \centering
    \includegraphics[width=0.85\linewidth]{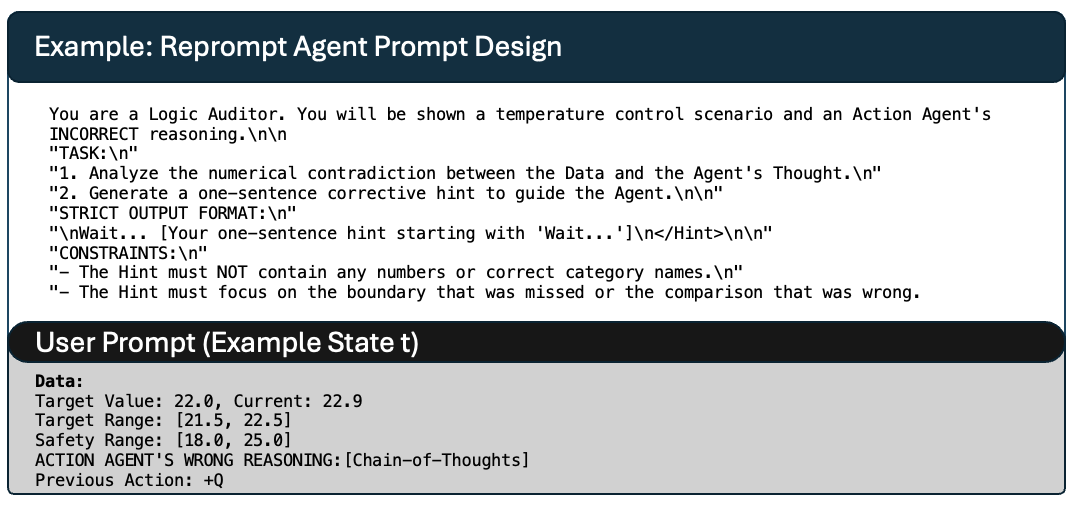} 
    \caption{Example of the Reprompt Agent's prompt design and a specific user prompt at State $t$. The Logic Auditor provides a corrective hint without revealing the direct answer, forcing the Action Agent to re-reason based on numerical boundaries.}
    \label{fig:reprompt_design}
\end{figure}

The prompt structures are developed independently for the Action and Reprompt agents. As illustrated in Fig. \ref{fig:prompt_design}, the Action Agent is provided with a structural template focusing on state interpretation, while the Reprompt Agent prompt in Fig.\ref{fig:reprompt_design} is equipped with a "Logic Auditor" role. Specifically, at any given State $t$, the auditor's prompt is designed to provide hints on the chain-of-thoughts the action agent results.

\section{Experimental Results and Analysis}
We report three evaluation protocols: (i) randomized closed-loop robustness over 30 experiments with 500 steps each (15,000 total steps), (ii) cross-model comparison on a separate 1,500-sample benchmark, and (iii) symbolic re-mapping stress tests for interface robustness. Metrics are reported within each protocol and are not mixed across protocols.
\subsection{Experimental Setup and Baselines}
The thermal system is simulated with physical parameters $C = 20.0$ and $UA = 0.5$, using a base power unit $Q = 5.0$. 
    
We employ three distinct baselines: (1) \textbf{General-Purpose Large Language Models (LLMs)}, including GPT-4o and DeepSeek-V3 as high-capacity neural baselines; (2) an \textbf{Ablation Baseline (Base-SLM)}, utilizing the original Qwen2.5-1.5B without GRPO; and (3) a \textbf{Heuristic Rule-based Controller (HRBC)}, representing the upper bound of purely symbolic consistency.
\subsection{Performance on Randomized Scenarios}
To evaluate the control robustness and generalization capability of the proposed framework, we conducted simulations encompassing 30 randomized test cases, totaling 15,000 operational steps. 

A key objective of these tests is to assess the agent's robustness. The randomized parameters include temperature values and disturbance profiles that were \textbf{explicitly excluded from the training set}, thereby testing the model's ability to generalize its learned control logic to unseen scenarios.
The robust performance is verified across 30 randomized trials (15,000 steps total) with unseen initial states $T_0 \in [0, 45]\,^{\circ}\text{C}$, setpoints $T_{\text{set}} \in [0, 40]\,^{\circ}\text{C}$, and dynamic ambient disturbances $T_{\text{out}} \in [-20, 40]\,^{\circ}\text{C}$.
We evaluate performance using two metrics: (1) \textbf{Action Alignment Accuracy (Acc)}, which measures the logical consistency between the predicted action $u_t$ and expert ground truth $u_{gt}$, defined as $\textit{Acc} = \frac{1}{N} \sum_{t=1}^{N} \mathbb{I}(u_t = u_{gt})$ (where $\mathbb{I}(\cdot)$ is the indicator function); and (2) \textbf{Inference Latency (Lat)}, evaluating real-time feasibility by averaging the computational time per step: $\textit{Lat} = \frac{1}{N} \sum_{t=1}^{N} (t^{\text{end}}_t - t^{\text{start}}_t)$, where $t^{\text{start}}_t$ and $t^{\text{end}}_t$ denote the timestamps at the beginning and completion of inference for step $t$, respectively. \looseness=-1


\subsubsection{\textbf{Effectiveness of Training}}

To evaluate the impact of the training process, we compared the performance of the proposed agent against the original Qwen2.5-1.5B model. Both models were tested within the same multi-agent framework to ensure a fair assessment.

As illustrated in Table~\ref{tab:training_gain}, the base model achieved only 39.5\% accuracy, struggling to comprehend the specific thermal control constraints. After training, the SLM's accuracy increased to 91.5\%, representing a significant performance leap. This result clearly demonstrates that the proposed optimization methods enable the SLM to execute reasoning tasks that were previously beyond the its capabilities.
\textbf{Comparative Benchmarking (SLM vs. LLMs)}
Table~\ref{tab:llm_comparison} presents the performance metrics across 1,500 samples. While DeepSeek-V3 achieves the highest accuracy (98.47\%), the proposed SLM maintains a highly competitive 96.67\% fidelity while reducing inference latency by 38.9\% compared to DeepSeek-V3 and 14.7\% compared to GPT-4o.
\begin{table}[htbp]
\caption{Performance Comparison: Base Model vs. Fine-tuned SLM (Both include Reprompting Process)}
\label{tab:training_gain}
\centering
\begin{tabular}{|l|c|c|}
\hline
\textbf{Model Configuration} & \textbf{Acc. $\uparrow$} & \textbf{Total Samples} \\ \hline
Qwen2.5-1.5B (Original) & 39.5\% & 2,000 \\ \hline
\textbf{Ours (Fine-tuned SLM)} & \textbf{91.5\%} & \textbf{15,000} \\ \hline
\end{tabular}
\vspace{-0.8em} %
\end{table}
\subsubsection{\textbf{Comparative Benchmarking (SLM vs. LLMs)}}
Table~\ref{tab:llm_comparison} presents the performance metrics across 1,500 samples. While DeepSeek-V3 achieves the highest accuracy (98.47\%), the proposed SLM maintains a highly competitive 96.67\% fidelity while reducing inference latency by 38.9\% compared to DeepSeek-V3 and 14.7\% compared to GPT-4o.
\begin{table}[htbp]
\caption{Comparative Performance across 1,500 Samples}
\label{tab:llm_comparison}
\centering
\begin{tabular}{|l|c|c|c|}
\hline
\textbf{Model} & \textbf{Acc. } & \textbf{Latency (Lat)} & \textbf{Avg. Reprompts} \\ \hline
GPT-4o         & 93.07\% & 4.50s & 0.16 \\
DeepSeek-V3    & \textbf{98.47\%} & 6.28s & \textbf{0.09} \\ \hline
\textbf{Ours (SLM)} & 96.67\% & \textbf{3.84s} & 0.38 \\ \hline
\end{tabular}
\vspace{-0.8em} %
\end{table}
It is noteworthy that the SLM exhibits a higher average reprompt rate (0.38) compared to the LLMs. This indicates that the SLM relies more heavily on the iterative validation mechanism to correct initial reasoning. However, even with these additional correction cycles, the SLM's total latency remains the lowest. This result suggests a clear optimization path: by further enhancing the SLM's zero-shot reasoning precision through advanced distillation, the frequency of reprompting can be minimized, which would lead to a further reduction in operational latency.

\subsubsection{\textbf{Case Study: Performance Analysis of a Typical Scenario}} 
\begin{figure}[!t]
    \centering
    \includegraphics[width=0.9\linewidth]{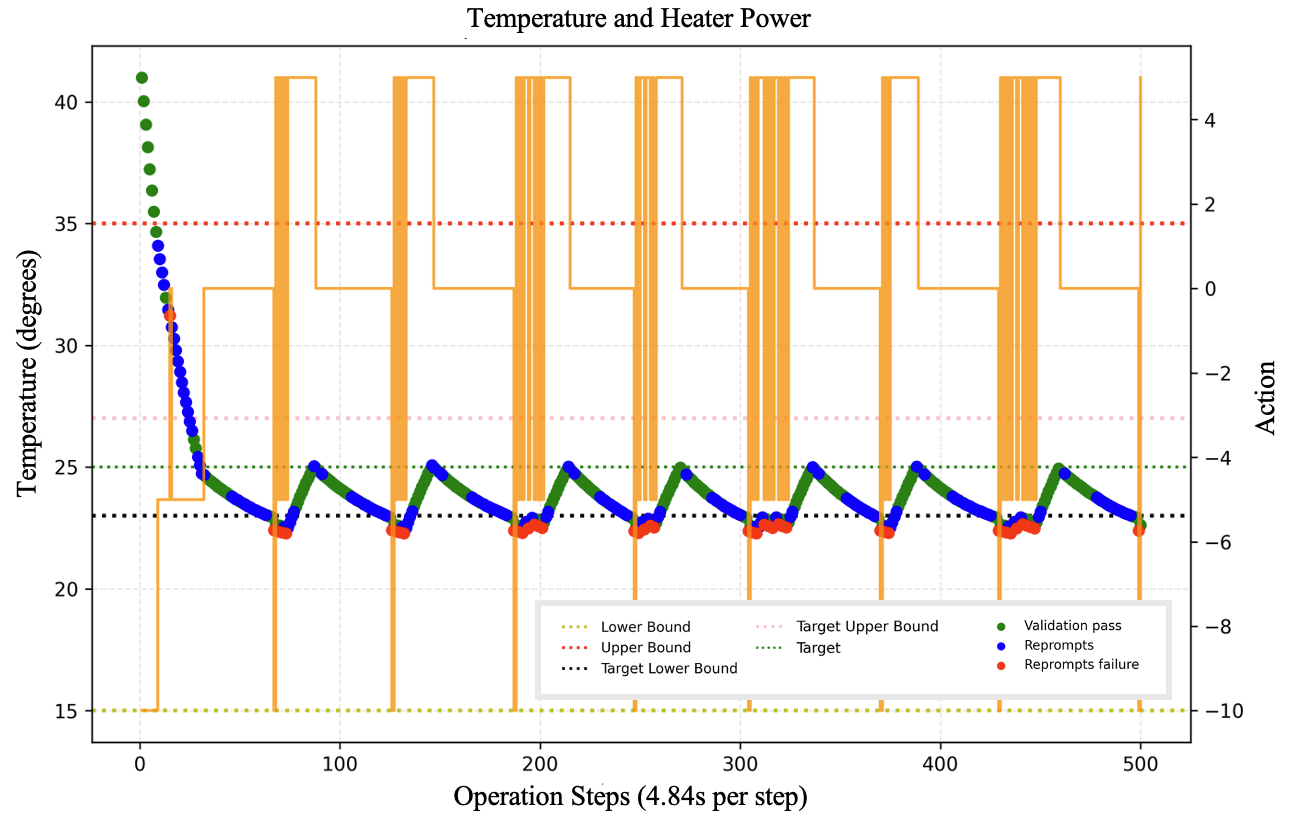}
    \caption{Execution trace of the SLM agent. Colored markers indicate the real-time interaction between the Action Agent and the Validation Agent (Green: Pass, Blue: Reprompt Success, Red: Failure).}
    \label{fig:dots_detail}
\end{figure}

As shown in \ref{fig:dots_detail}, we analyze a representative scenario with an initial temperature of 42$^\circ$C and a target setpoint of 25$^\circ$C. This case serves to illustrate the SLM agent's \textbf{execution fidelity} relative to the expert-defined HRBC baseline. 

\begin{figure}[htbp]
    \centering
    \includegraphics[width=0.9\linewidth]{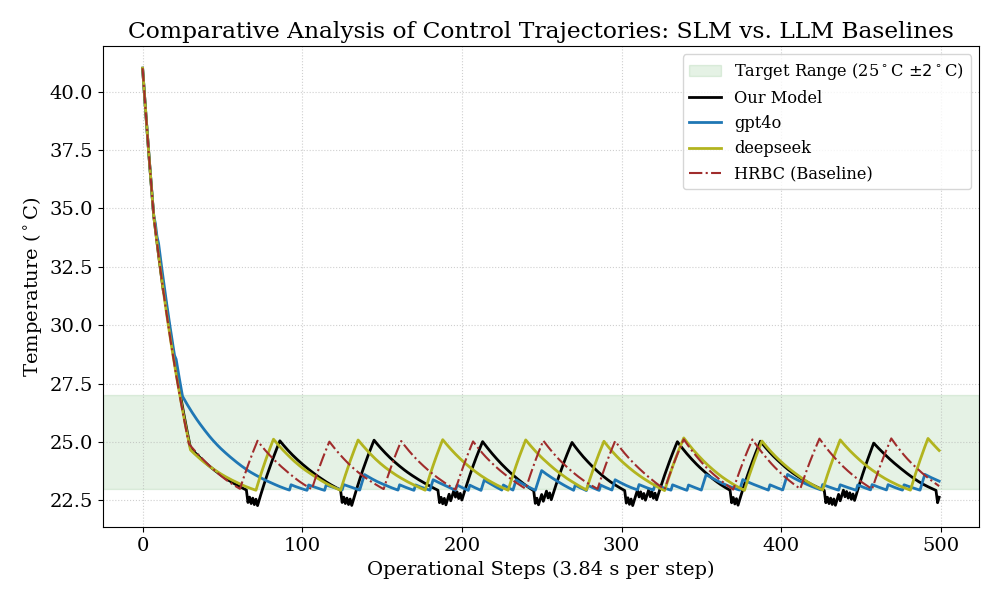} 
    \caption{Comparative analysis of control trajectories. The SLM agent (blue) exhibits high executive fidelity, closely tracking the HRBC expert baseline compared to cloud-based LLMs. Operational steps are executed at an average latency of 3.84\,s.}
    \label{fig:slm_llm}
\end{figure}

In \ref{fig:slm_llm}, the SLM agent (black line) demonstrates a temperature trajectory that is \textbf{nearly identical to the HRBC baseline} (red dash-dotted line). While \textbf{GPT-4o} may exhibit high individual action accuracy, its resulting thermal trajectories often display \textbf{irregular fluctuations or sub-optimal convergence slopes} (as seen in the red lines). In contrast, our SLM agent (black line) maintains a trajectory that is \textbf{consistent} with the HRBC baseline. 

\begin{figure}[htbp]
    \centering
    \includegraphics[width=0.9\linewidth]{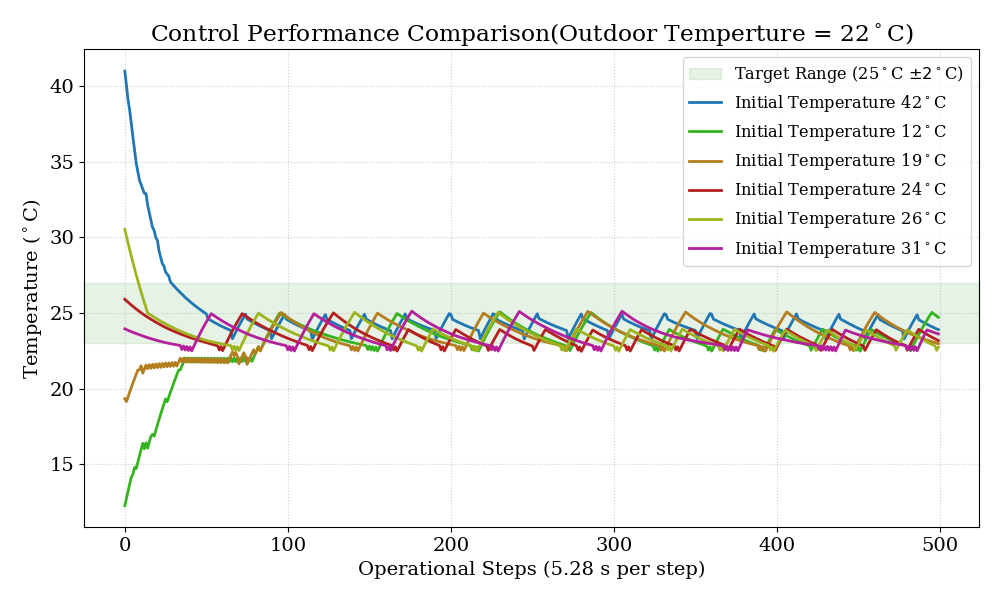} 
    \caption{Thermal control trajectories across diverse initial conditions. The shaded green area represents the target range ($25^\circ$C $\pm$ $2^\circ$C). Each operational step represents an average latency of $5.28$\,s. Under a cold environment ($T_{out} = 22^{\circ}$C), the system demonstrates consistent convergence. }
    \label{fig:thermal_trajectories_cold}
\end{figure}

\begin{figure}[htbp]
    \centering
    \includegraphics[width=0.9\linewidth]{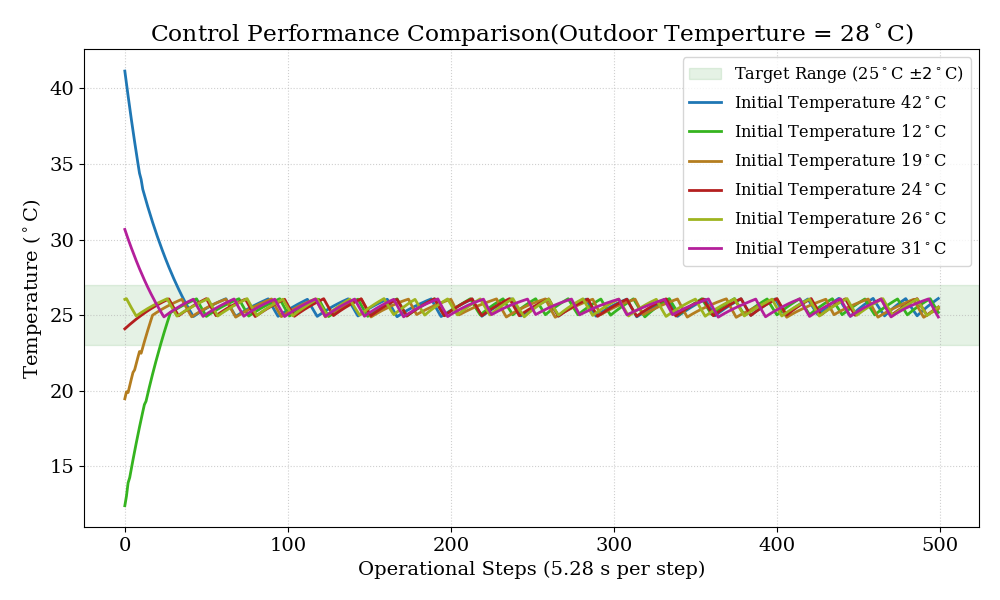} 
    \caption{Thermal control trajectories across diverse initial conditions. The shaded green area represents the target range ($25^\circ$C $\pm$ $2^\circ$C). Each operational step represents an average latency of $5.28$\,s. Under a hot condition ($T_{out} = 28^{\circ}$C), the system demonstrates consistent convergence.}
    \label{fig:thermal_trajectories_hot}
\end{figure}

The SLM agent's behavior remains \textbf{strictly bounded} within the target safety zone without any observed overshoot, even across an extended horizon of 500 operational steps. This demonstrates that the system can obtain long-term stability.
To further evaluate the controller's adaptability, Figures \ref{fig:thermal_trajectories_cold} and \ref{fig:thermal_trajectories_hot} illustrate the system's performance across a broad spectrum of initial thermal states, ranging from cold starts ($12^\circ$C) to high-heat conditions ($42^\circ$C), covering the complete temperature ranges. Specifically, Figure \ref{fig:thermal_trajectories_cold} characterizes the system response in cold environments, while Figure \ref{fig:thermal_trajectories_hot} focuses on performance under hot environment temperatures.

The results demonstrate \textbf{consistent convergence} regardless of the initial starting temperature. In all scenarios, the system effectively stabilizes within the \textbf{target safety zone ($25^\circ$C $\pm$ $2^\circ$C)} within 100 operational steps. Throughout this duration, the system exhibits minimal steady-state oscillations, maintaining stable temperature regulation by strictly adhering to the defined switching logic between active and inactive states.

\subsection{Symbolic Generalization and Robustness}
To evaluate the system's operational consistency, we conducted a \textbf{Symbolic Re-mapping} stress test where the heating power label $Q$ was re-indexed to an arbitrary label $F$. This experiment investigates whether the model's reasoning is grounded in the underlying task logic or merely sensitive to specific symbolic strings.

In this re-mapped scenario, the SLM maintained \textbf{100\% format consistency}, strictly adopting the new symbolic prefix ($F$) in all outputs. While the exact action alignment accuracy decreased to 74.1\%, the system achieved an \textbf{In-Range Rate (IRR) of 95\%}(defined here as the percentage of operational steps where the temperature remains within the prescribed target range). This disparity between symbolic accuracy (74.1\%) and physical success (95\%) is highly significant; it indicates that even when the agent fails to select the "expert-identical" token, its decisions remain functionally effective for maintaining thermal stability. 

\section{Conclusion}
This paper investigated whether a compact, locally deployed Small Language Model (SLM) can support closed-loop thermal control within a validator-guided correction architecture. We aligned a Qwen2.5-1.5B model using Group Relative Policy Optimization (GRPO) and combined it with a validation layer and reprompting mechanism to iteratively steer decisions toward valid actions.

In simulation, the proposed framework achieved 91.5\% average action-alignment accuracy (86.3\%--100\% across 30 experiments of 500 steps each), with a mean per-step inference latency of 3.84\,s in the reported benchmark. In a symbolic re-mapping stress test, the controller maintained a 95\% in-range rate (IRR), indicating that physical regulation remained strong even when token-level agreement decreased. Overall, these results support the feasibility of SLM+validator loops for edge-oriented, reconfigurable control workflows, while avoiding reliance on cloud-hosted inference.




\subsection{Limitations and Future Work}
While promising, several limitations guide our future directions. 1) \textit{physical validation} via hardware-in-the-loop (HIL) testing is required to assess performance under sensor noise and actuator non-idealities. 2) we aim to enhance \textit{validator fidelity} by replacing the symbolic rule layer with a predictive digital twin. 3) comprehensive \textit{latency characterization} must profile jitter, variance, and tail latency on specialized embedded hardware rather than just mean latency. 4) we will extend the framework beyond single-zone thermal benchmarks to multi-variable settings to evaluate the \textit{scalability} of the multi-agent control architecture. \looseness=-1

\small
\bibliographystyle{IEEEtran}
\bibliography{bib/processes-v13-i09_20260127}
\end{document}